\pdfoutput=1

\documentclass[11pt]{article}

\usepackage[]{ACL2023}

\usepackage{times}
\usepackage{latexsym}

\usepackage{booktabs}
\usepackage{multirow}
\usepackage{url}
\usepackage{times}
\usepackage{latexsym}
\usepackage{graphicx}
\usepackage{array}
\usepackage{multicol}
\usepackage{enumitem}
\usepackage{caption}

\usepackage[T1]{fontenc}

\usepackage[utf8]{inputenc}

\usepackage{microtype}

\usepackage{inconsolata}

\title{An Open-Source Data Contamination Report for Large Language Models}

\author{Yucheng Li\textsuperscript{1}\space, Frank Guerin\textsuperscript{1}\space, Chenghua Lin\textsuperscript{2}\space \\
\addlinespace[0.3em]
\textsuperscript{1}~University of Surrey, UK  \quad
\textsuperscript{2}~University of Manchester, UK \\
\texttt{\{yucheng.li, f.guerin\}@surrey.ac.uk}\\
\texttt{chenghua.lin@manchester.ac.uk}
}

\begin{document}
\maketitle
\begin{abstract}
Data contamination in model evaluation has become increasingly prevalent with the growing popularity of large language models. It allows models to "cheat" via memorisation instead of displaying true capabilities. Therefore, contamination analysis has become an crucial part of reliable model evaluation to validate results. However, existing contamination analysis is usually conducted internally by large language model developers and often lacks transparency and completeness. This paper presents an extensive data contamination report for over 15 popular large language models across six popular multiple-choice QA benchmarks. We also introduce an open-source pipeline that enables the community to perform contamination analysis on customised data and models. Our experiments reveal varying contamination levels ranging from 1\% to 45\% across benchmarks, with the contamination degree increasing rapidly over time. Performance analysis of large language models indicates that data contamination does not necessarily lead to increased model metrics: while significant accuracy boosts of up to 14\% and 7\% are observed on contaminated C-Eval and Hellaswag benchmarks, only a minimal increase is noted on contaminated MMLU. We also find larger models seem able to gain more advantages than smaller models on contaminated test sets.
\end{abstract}

\section{Introduction}

Recent years have seen remarkable progress in language models pre-trained on massive text corpora scraped from the web. However, many widely used evaluation benchmarks are also constructed from similar web sources, leading to a concerning issue of \textit{data contamination} where examples from test sets are unintentionally included in training data. Contamination enables models to "cheat" via memorisation of test data rather than displaying true generalisation \cite{marie2023}, which creates an illusion of progress, distorts model comparisons, and undermines the utility of benchmarks \cite{jacovi2023stop,sainz-etal-2023-nlp}.

Contamination analysis therefore became a crucial part of reliable LLM evaluation to validate the results. However, existing contamination analysis is often conducted internally by LLM developers and often lacks transparency and completeness. For instance, OpenAI’s contamination study for GPT-4 \cite{openai2023gpt4} only covered the pre-training data and omitted later fine-tuning stages. Llama-2 \cite{touvron2023llama2} only reported contamination statistics for 2 of the 20+ benchmarks used in their evaluation. In addition, their implementation details of contamination identification remains unclear. Overall, existing internal contamination studies tend to lack sufficient transparency, with minimal sharing of comprehensive contamination measurements across all evaluation benchmarks, as well as training data details and code to reproduce the results. This prevents the wider research community from fully auditing the credibility of reported metrics and model capabilities.

This paper presents an open contamination analysis for over 15 popular large language models on six common multiple-choice benchmarks, aiming to provide more comprehensive measurements and insights compared to limited existing studies. The analysis includes a range of foundation models such as LLaMA \cite{touvron2023llama}, Llama-2, Yi \cite{Yi}, Mistral \cite{jiang2023mistral}, Baichuan \cite{yang2023baichuan}, and Qwen \cite{bai2023qwen} across multiple model sizes (7B, 13B, 30B, 34B, 65B, 70B parameters) as well as instruct-tuned models built on these foundations like Llama-2 Chat and Mistral-Instruct. Six widely used multi-choice benchmarks are assessed: Winogrande \cite{sakaguchi2021winogrande}, AI2\_ARC \cite{Clark2018ThinkYH}, CommonsenseQA \cite{talmor2018commonsenseqa}, HellaSwag \cite{zellers2019hellaswag}, MMLU \cite{hendryckstest2021}, and C-Eval \cite{huang2023ceval}. Our methodology proceeds in four steps: first, we verify whether test examples appear in Common Crawl\footnote{\url{https://commoncrawl.org/}}, a popular large corpus often used in language model pre-training. If a test example is found verbatim in Common Crawl, it was very likely included in the pre-training phrase of language models, making it a "contaminated" sample. Based on the presence of test samples, we then categorise benchmarks into the clean set and contaminated set. Finally, we compare model performance on these subsets to assess the impact of data contamination on evaluation results. At the end of the paper, we compare our analysis to Llama-2's original contamination results and discuss the effectiveness of existing contamination mitigation methods.

Our analysis reveals the following key findings: 1) we detect varying levels of data contamination across benchmarks, with 1\% to 45.8\% of examples showing verbatim overlap with Common Crawl; 2) by comparing the contamination degree between Common Crawl Dec 2020 to Oct 2023, we find data contamination grows rapidly through time; 3) data contamination does not necessarily lead to increased model performance: we found significant accuracy boosts of 14\% and 7\% on C-Eval and Hellaswag, but very little increase on MMLU; 4) we also find a tendency that larger models seems to obtain more advantages than smaller models from data contamination, perhaps due to the more powerful memorisation capacities of larger models; 5) finally, we show our results align well with  Llama's original contamination reports, demonstrating the effectiveness of our method.

\section{Data Contamination}
\label{background}

\noindent\textbf{What is data contamination?} Data contamination refers to the phenomenon that examples from the test set are also found in the training data. This might lead to the evaluation failing to accurately reflect models' capabilities, as models can cheat by memorising instead of learning to generalise. There are two primary types of data contamination \cite{dodge2021documenting}: \textit{input contamination} refers to only the input appearing in the pretraining corpus, and \textit{input-and-label contamination} is when both inputs and their labels are present. The latter is generally more problematic, as models can directly memorise input-output pairs. But the first can still cause issues as models may gain an advantage even if only the input is learned (see \S\ref{impact} for details), especially for assessing few-shot and zero-shot learning capabilities.

\noindent\textbf{How common is data contamination?} Data contamination appears to be quite widespread across commonly used NLP benchmark datasets based on findings from recent studies. \citet{dodge2021documenting} revealed exact match contamination rates ranging from under 2\% to over 50\% on various GLUE benchmarks when compared to the C4 pretraining data. The GPT-3 study \cite{brown2020language} found over 90\% of examples in Quac, SQuADv2, and DROP were flagged as contaminated. FLAN \cite{wei2021finetuned} evaluations identified 7 out of 26 datasets exhibiting a serious contamination ratio of 50\% and over. Llama-2 \cite{touvron2023llama} reported over 16\% of MMLU examples are contaminated and about 11\% are seriously contaminated (more than 80\% token leakage). GPT-4 \cite{openai2023gpt4} use academic exams instead of NLP benchmarks for model evaluation. While 4 out of 34 exams are found have zero contamination (e.g., Leetcode and Bar Exam), 9 out of 34 showed over 20\% of instances are marked as dirty examples. In summary, we found data contamination is becoming an increasingly prevalent issue for LLMs, which must be carefully measured and accounted for in order to accurately assess model performance.

\noindent\textbf{How to identify data contamination?} \citet{dodge2021documenting} takes a straightforward approach to detect exact matches between test set examples and the pretraining data after normalising for capitalisation and punctuation. The \textit{exact match here} means the entire input of an evaluation text is found in the training data. The GPT-3 paper \cite{brown2020language} uses n-gram overlap to identify contamination, treating any examples with 13-gram co-occurrence in both test sets and training data as dirty examples. PaLM \cite{chowdhery2022palm} considers a sample to be contaminated if 70\% of its 8-grams can be found at least once in the training data. Llama-2 matches on verbalised and tokenized input to allow a token-level approach to identify contamination. It also involves a "skipgram budget" to allow slight variants in overlapping. Overall, existing approaches usually use substring matching between evaluation examples and training data to identify data contamination. However, if we have no access to the training data, which is often the case for most recent closed models, it is extremely difficult to reveal contamination by observing models themselves.  In this paper, we utilise a pipeline consisting of a search engine and Common Crawl index for detecting potentially contaminated test samples, avoiding the need for full training data. This enables the community and third parties to conduct contamination analysis.

\noindent\textbf{To what extent does data contamination affect model evaluation?} While contaminated data can potentially inflate scores, models do not necessarily perform worse on clean subsets or better on dirty subsets across all datasets. The degree of impact likely depends on factors like the dataset characteristics, model scale, and nature of the pre-training data. For instance, GPT-3 \cite{brown2020language} showed a small 1-2\% performance drop on clean subsets for PIQA and ReCoRD, comparing to a significant 6\% drop on clean set of SQuAD as 94\% of its test examples were contaminated. The LLaMA model \cite{touvron2023llama} did not show significant gaps between clean and dirty subset performance. On HellaSwag, LLaMA's 70B model showed a 15.3 point gap between clean (63.5) and dirty (78.8) subsets. Detecting and accounting for data contamination remains an active area of research, as there is no consensus yet on best methodologies and acceptable contamination levels.

\section{Benchmarks for Language Models}

Clean and robust benchmarks are the key to guide further progress of various models in NLP. 
Popular benchmarks used to evaluate large language models include:

\begin{quote}
\noindent \textbf{Comprehensive}: MMLU, Big Bench \cite{srivastava2023imitation}, AGI Eval \cite{zhong2023agieval}, C-Eval

\noindent \textbf{Commonsense reasoning}: PIQA \cite{bisk2019piqa}, SIQA \cite{sap2019socialiqa}, HellaSwag, WinoGrande, ARC, OpenBookQA \cite{Mihaylov2018CanAS}, CommonsenseQA

\noindent \textbf{World knowledge}: NaturalQuestions \cite{kwiatkowski2019natural}, TriviaQA \cite{joshi2017triviaqa}

\noindent \textbf{Reading comprehension}: SQuAD \cite{rajpurkar2018know}, QuAC \cite{choi2018quac}, BoolQ \cite{clark2019boolq}

\noindent \textbf{Math}: GSM8K \cite{cobbe2021training}, MATH \cite{hendrycks2021measuring}

\noindent \textbf{Code}: HumanEval \cite{chen2021humanevaluating}, MBPP \cite{austin2021mbpp}
\end{quote}
As many of their construction rely heavily on online materials, they are highly prone to data contamination as their source spread on the internet. In this paper, we analyse six representative multi-choice QA benchmarks: MMLU, C-Eval, Winogrande, CommonsenseQA, ARC, and Hellaswag. These benchmarks have been selected due to their varied sources and potential susceptibility to data contamination. MMLU, ARC, and C-Eval, which are academic test-based benchmarks, were compiled from online \texttt{.docx/.pdf} files using techniques like OCR, typically assumed to be less affected by data contamination as such files are often not indexed by online crawlers. However, C-Eval stands out as it is a non-English (Chinese) benchmark, offering an opportunity to assess the impact of non-English benchmarks on language models. Winogrande, uniquely human-authored from scratch, allows examination of whether manually created benchmarks are less prone to data contamination. CommonsenseQA and Hellaswag, both internet-sourced, differ in their source popularity; while CommonsenseQA is built upon the less influential ConceptNet, Hellaswag is sourced from the more popular WikiHow. This selection of benchmarks provides a comprehensive overview of how different sourcing and construction methods might influence the presence and extent of data contamination in language model evaluations.

\section{Our Approach}

\begin{figure*}
    \centering
    \includegraphics[width=0.8\textwidth]{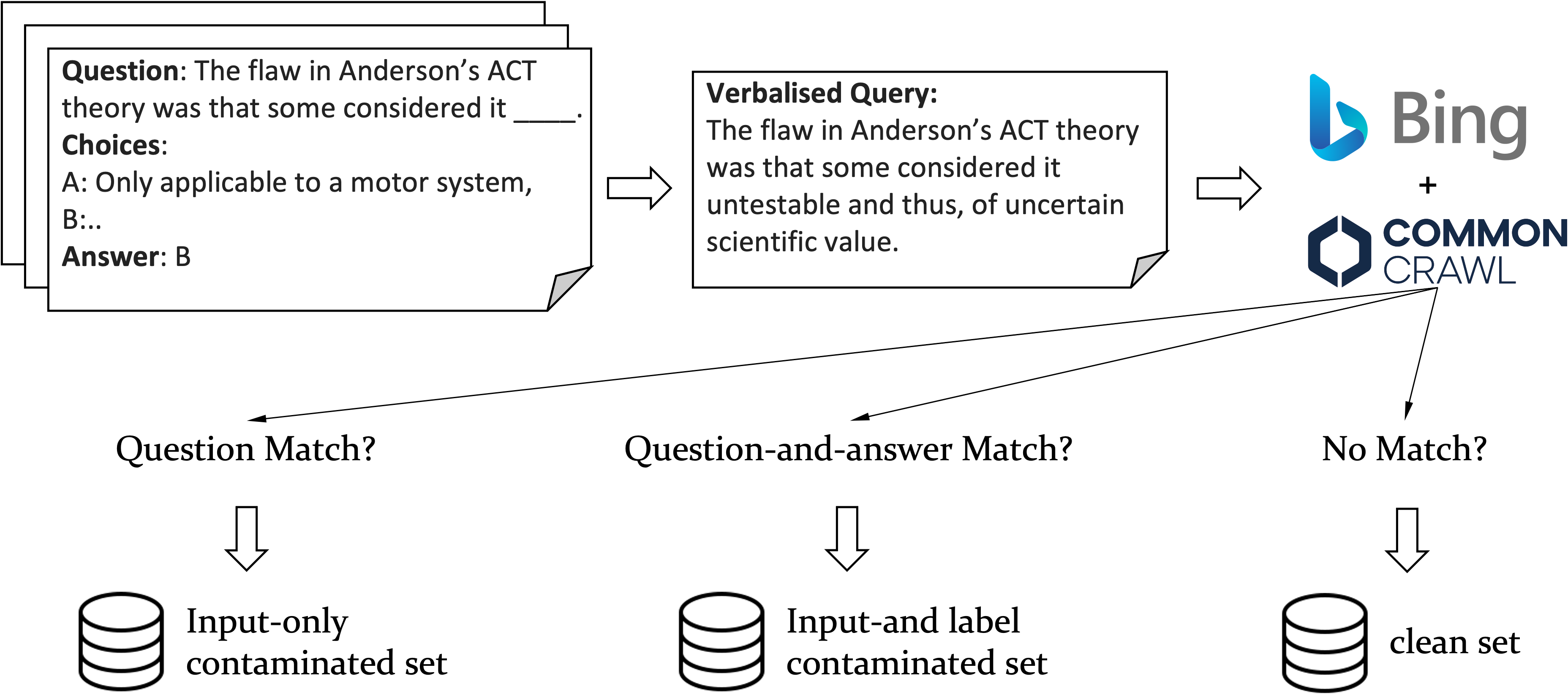}
    \caption{The categorisation of contaminated test samples.}
    \label{fig:enter-label}
\end{figure*}

The central goal of data contamination analysis is to categorise test samples as either clean or contaminated and then evaluate models separately on the clean and contaminated samples to assess the impact of contamination on the performance metrics. In this section, we describe our methodology to identify contaminated test samples.
The basic idea of detecting contaminated examples in our method is to check whether test examples appear verbatim in Common Crawl. We base on Common Crawl because it is completely open-sourced and often comprises the majority of pre-training data for large language models, e.g., Common Crawl weights over 80\% in GPT-3 and LLaMA training data \cite{brown2020language, touvron2023llama}. 

We tailor our search window based on each model's training data collection period. For example, LLaMA models use Common Crawl dumps from 2017 to 2020, which makes our contamination search window 2017-2020. For models with unknown data ranges, we use a period from 2017 to the release date as the search window. For example, Mistral 7B was released in Oct 2023 without details on their training data range. So we use a search window of Jan 2017 to Oct 2023.

Since Common Crawl produces 20TB of data monthly, exhaustively searching the full contents is infeasible. Therefore, we try to find overlap between benchmarks and training data in two steps: First, 1) we use the Bing Search API to check if verbatim test examples appear online, which likely indicates inclusion in Common Crawl. Second, 2) we specifically verify if pages containing verbatim test examples were indexed in Common Crawl, by only searching the URLs rather than full contents. This reduces the disk usage to a manageable 2.5TB for detecting training data overlap. The \textit{freshness} parameter in the Bing API is configured to align the search window with the expected training data range of each model.

To construct the search queries, we verbalise examples accordingly and make sure the question and the correct answer are involved in the queries. For example:

\begin{quote}
    \textbf{Question}: The flaw in Anderson’s ACT theory was that some considered it \_\_\_\_.
    
    \textbf{Choices}: 

    A: 'Only applicable to a motor system',
    
    B: 'Untestable and thus, of uncertain scientific value',
    
    C: 'Lacking in definition for its elements',
    
    D: 'Overly complex in explaining the operation of cognition',

    \textbf{Answer}: B

    \textbf{\textit{Verbalised Query}}: The flaw in Anderson’s ACT theory was that some considered it untestable and thus, of uncertain scientific value.
\end{quote}
We verbalise this multi-choice question to a query by filling the correct answer to the blank. We do not include other options in the query, because, as discussed in section \ref{background}, the presence of other options does not matter. The question and answer are the key for identifying data contamination. If there is no blank in the question, we simply append the answer after the question to form the query.

\begin{table*}[t]
    \centering
        \resizebox{0.9\textwidth}{!}{%
    \begin{tabular}{lrrr>{\raggedleft\arraybackslash}p{2.8cm}>{\raggedleft\arraybackslash}p{2.5cm}>{\raggedleft\arraybackslash}p{2.8cm}}
    \toprule
    \textbf{Dataset} & \textbf{Split} & \textbf{\#Total} & \textbf{\#Online} & \textbf{\#All Dirty (in CommonCrawl)} & \textbf{\#Input-only Contamination} & \textbf{\#Input-and-label Contamination} \\
    \midrule
    ARC\_c & Test & 1172 & 372 & 336 (28.7\%) & 53 (4.5\%) & 283 (24.1\%) \\
    CommonsenseQA & Dev & 1221 & 44 & 20 (1.6\%) & 3 (0.2\%) & 17 (1.4\%) \\
    Winogrande & Dev & 1267 & 54 & 14 (1.1\%) & 0 (0.0\%) & 14 (1.1\%) \\
    C-Eval & Dev & 1346 & 618 & 616 (45.8\%) & 69 (5.1\%) & 547 (40.6\%) \\
    Hellaswag & Dev & 10042 & 1690 & 1247 (12.4\%) & 46 (0.4\%) & 1201 (12.0\%) \\
    MMLU & Test & 13987 & 4285 & 4077 (29.1\%) & 678 (4.8\%) & 3399 (24.3\%) \\
    \bottomrule
    \end{tabular}}
    \caption{Data contamination statistics for multi-choice QA benchmarks. Search window: 2020.10-2023.10.}
    \label{tab:contamination}
\end{table*}

To identify overlapping between test samples and training data, existing methods often rely on exact string matches. For example, \citet{brown2020language} use N-gram overlapping ranging from 8-grams to maximum 13-grams for all evaluation tasks. GPT-4's criterion for contamination is sub-string matching with at least 50 characters \cite{openai2023gpt4}. However, according on our manual analysis, we find the approach of exact string matches often lead to false negative in our pipeline. \citet{touvron2023llama2} propose a more fine-grained method that assesses contamination in token-level and involves a small "skipgram budget" to accommodate slight variations of sequences. However, their exact implementation details remain unclear.
We instead simply compute METEOR \cite{banerjee2005meteor} score between matched pages and the queries to quantify the extent of oerlap. We consider examples with a METEOR recall score over 0.75 as contaminated cases. This method tolerates minor inserted phrases and word form variations, which greatly mitigates false negative issue that strict string matching would miss. To avoid potential false positives, we configure our method with two key settings: 1) an order penalty (gamma of 0.8) for METEOR ensures matches respect sequence; 2) matching is constrained to a window up to 2x the query length, preventing partial or out-of-context matches. We compare our approach against to Llama-2's in section \ref{compare to llama}.

According to section \ref{background}, here we distinguish two types of data contamination: 1) \textit{input contamination} where only question is presented in the matched pages but not answer; 2) \textit{input-and-label contamination} where both question and answer occur in the matched pages. In the upcoming sections, these two types of data contamination are compared and analysed separately.

\section{Contamination Statistics for Multi-Choice Benchmarks}
\label{contamination_statistics}

Our analysis reveals varying levels of data contamination across six multi-choice QA benchmarks, as shown in Table \ref{tab:contamination}. According to the table, we have the following key findings. First, Academic test-based benchmarks like MMLU and C-Eval, despite being collected through methods like OCR, exhibit the highest levels of contamination (29.1\% and 45.8\%, respectively). This high rate is attributed to the widespread distribution and communication of academic test examples, making them more prone to sharing and discussion. In contrast, benchmarks manually created from scratch like Winogrande demonstrate minimal contamination (1.1\%), as they avoided using internet resources in their benchmark construction. Third, we find significant differences among internet-sourced benchmarks. For example, CommonsenseQA has low contamination (1.6\%) but HellaSwag has a rather significant higher overlap (12.4\%). This variation might stem from different popularity of the sources: ConceptNet, as the source of CommonsenseQA is less popular than WikiHow, the source of HellaSwag. Finally, we find most contamination belongs to \textit{input-and-label} contamination, indicating models can often find the answer alongside with the question for contaminated test samples.

\begin{figure}
    \centering
    \includegraphics[width=\columnwidth]{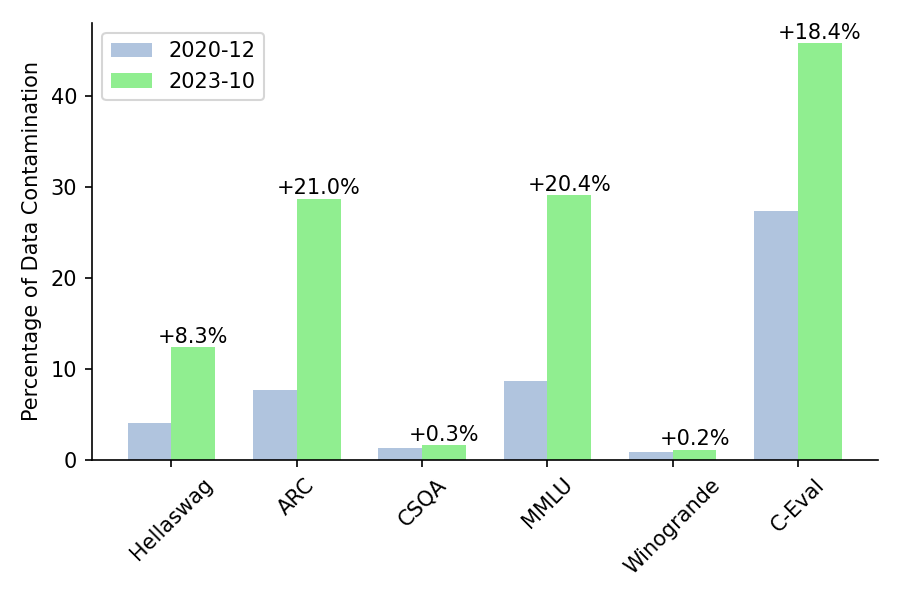}
    \caption{Increase in Data Contamination from \texttt{2017-2020} to \texttt{2020-2023}. CSQA stands for CommonsenseQA.}
    \label{fig:increase}
\end{figure}

\begin{figure}[t]
    \centering
    \includegraphics[width=0.8\columnwidth]{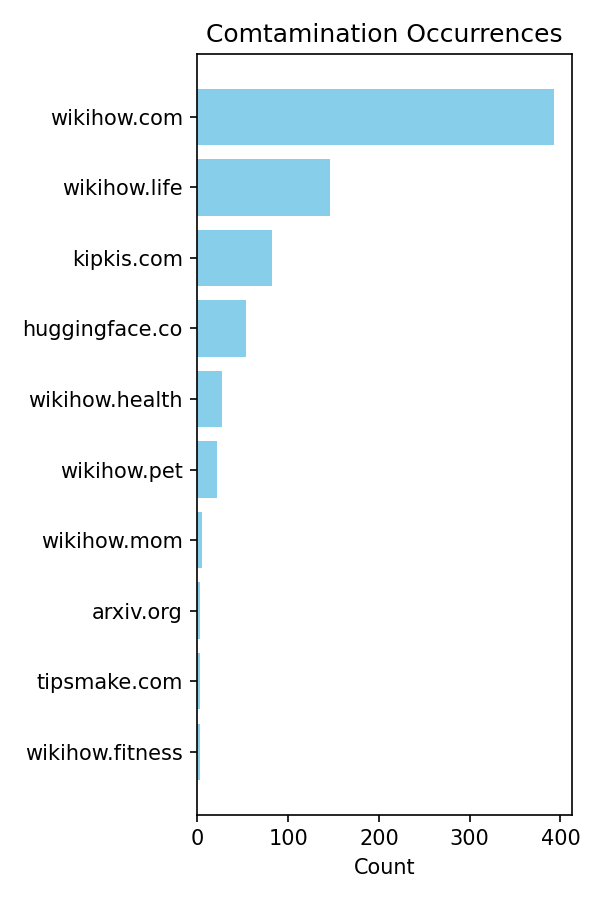}
    \caption{Domains analysis for data contamination in Hellaswag.}
    \label{fig:contamination-distribution}
\end{figure}

We also illustrate how data contamination increase over time, as shown in Figure \ref{fig:increase}. In the figure, benchmarks such as CommonsenseQA and Winogrande maintain very low rates of contaminated data, with increases of just 0.3\% and 0.2\% over the past three years. However, benchmarks collected from academic tests like ARC, MMLU, and C-Eval have experienced substantial increase in contamination, with up to 21\% of examples flaged as contaminated during the same period. This shows how test content in academic benchmarks can easily propagate across the internet, which can be a serious issue for academic test based language model benchmarks. We also observe a moderate 8.3\% increase for Hellaswag, further demonstrating the increasing risk of data contamination for internet sourced benchmarks.

In Figure \ref{fig:contamination-distribution}, we illustrate where these Hellaswag contaminated test samples come from. We discover that data contamination manifests in a centralised fashion, which means contaminated test samples are not evenly distributed across domains. Instead, they are significantly concentrated in specific domains and rare in others. This finding is meaningful as it reveals the possibility that blocking specific domains during training data collection might alleviate the issue of data contamination. You can find more domain analysis and contamination examples in Appendix \ref{more-info}.

\begin{table*}[t]
\centering
\resizebox{\textwidth}{!}{
\begin{tabular}{lrrrrrrrrrrrr}
\toprule
& \multicolumn{3}{c}{MMLU} & \multicolumn{3}{c}{Hellaswag} & \multicolumn{3}{c}{ARC} & \multicolumn{3}{c}{Average} \\
\cmidrule(lr){2-4}
\cmidrule(lr){5-7}
\cmidrule(lr){8-10}
\cmidrule(lr){11-13}
& Clean & Dirty \romannumeral 1. & Dirty \romannumeral 2. & Clean & Dirty \romannumeral 1. & Dirty \romannumeral 2. & Clean & Dirty \romannumeral 1. & Dirty \romannumeral 2. & Clean & Dirty \romannumeral 1. & Dirty \romannumeral 2. \\ \midrule
LLaMA 7B & .3518 & .3577 $\uparrow$ & .3488 $\downarrow$ & .6394 & .6696 $\uparrow$ & .6727 $\uparrow$ & .3627 & .3448 $\downarrow$ & .3167 $\downarrow$ & .4513 & .4574 $\uparrow$ & .4461 $\downarrow$ \\
LLaMA 13B & .4429 & .4309 $\downarrow$ & .4884 $\uparrow$ & .7073 & .7913 $\uparrow$ & .7909 $\uparrow$ & .3924 & .3563 $\downarrow$ & .3333 $\downarrow$ & .5142 & .5262 $\uparrow$ & .5375 $\uparrow$ \\
LLaMA 30B & .5381 & .5447 $\uparrow$ & .5930 $\uparrow$ & .7412 & .7913 $\uparrow$ & .7909 $\uparrow$ & .4249 & .4598 $\uparrow$ & .4667 $\uparrow$ & .5681 & .5986 $\uparrow$ & .6169 $\uparrow$ \\
LLaMA 65B & .6316 & .5447 $\downarrow$ & .6279 $\downarrow$ & .7613 & .8087 $\uparrow$ & .8091 $\uparrow$ & .4276 & .4713 $\uparrow$ & .4667 $\uparrow$ & .6068 & .6082 $\uparrow$ & .6346 $\uparrow$ \\ \midrule
Llama-2 7B & .4180 & .4309 $\uparrow$ & .4535 $\uparrow$ & .6746 & .7217 $\uparrow$ & .7182 $\uparrow$ & .3803 & .4368 $\uparrow$ & .4167 $\uparrow$ & .4910 & .5298 $\uparrow$ & .5295 $\uparrow$ \\
Llama-2 13B & .5596 & .5285 $\downarrow$ & .5814 $\uparrow$ & .8254 & .8087 $\downarrow$ & .8000 $\downarrow$ & .4221 & .4368 $\uparrow$ & .4167 $\downarrow$ & .6024 & .5913 $\downarrow$ & .5994 $\downarrow$ \\
Llama-2 70B & .6763 & .6667 $\downarrow$ & .7093 $\uparrow$ & .7726 & .8348 $\uparrow$ & .8455 $\uparrow$ & .4555 & .5632 $\uparrow$ & .5667 $\uparrow$ & .6348 & .6882 $\uparrow$ & .7072 $\uparrow$ \\ \midrule
Llama-2 Chat 7B & .4062 & .3851 $\downarrow$ & .4060 $\downarrow$ & .6760 & .7605 $\uparrow$ & .7632 $\uparrow$ & .3701 & .4474 $\uparrow$ & .5000 $\uparrow$ & .4841 & .5310 $\uparrow$ & .5564 $\uparrow$ \\
Llama-2 Chat 13B & .5417 & .5098 $\downarrow$ & .5279 $\downarrow$ & .7341 & .8055 $\uparrow$ & .8100 $\uparrow$ & .4334 & .5526 $\uparrow$ & .5769 $\uparrow$ & .5698 & .6226 $\uparrow$ & .6383 $\uparrow$ \\
Llama-2 Chat 70B & .6324 & .6159 $\downarrow$ & .6392 $\uparrow$ & .7576 & .8273 $\uparrow$ & .8341 $\uparrow$ & .4343 & .4737 $\uparrow$ & .4615 $\uparrow$ & .6081 & .6390 $\uparrow$ & .6450 $\uparrow$ \\ \bottomrule
Mistral 7B & .6501 & .6291 $\downarrow$ & .6463 $\downarrow$ & .8533 & .8287 $\downarrow$ & .8326 $\downarrow$ & .4720 & .5263 $\uparrow$ & .5769 $\uparrow$ & .6585 & .6614 $\uparrow$ & .6853 $\uparrow$ \\
Mistral-FT 7B & .5576 & .5403 $\downarrow$ & .5588 $\uparrow$ & .7168 & .6691 $\downarrow$ & .6727 $\downarrow$ & .4426 & .5000 $\uparrow$ & .5577 $\uparrow$ & .5723 & .5698 $\downarrow$ & .5964 $\uparrow$ \\
Yi 6B & .6481 & .6386 $\downarrow$ & .6551 $\uparrow$ & .7628 & .7533 $\downarrow$ & .7617 $\downarrow$ & .4380 & .4868 $\uparrow$ & .5000 $\uparrow$ & .6163 & .6262 $\uparrow$ & .6389 $\uparrow$ \\
Qwen 7B & .5785 & .5665 $\downarrow$ & .5825 $\uparrow$ & .9153 & .9144 $\downarrow$ & .9186 $\uparrow$ & .4096 & .4605 $\uparrow$ & .4423 $\uparrow$ & .6345 & .6471 $\uparrow$ & .6478 $\uparrow$ \\
Baichuan2 7B & .5594 & .5320 $\downarrow$ & .5491 $\downarrow$ & .7494 & .7199 $\downarrow$ & .7240 $\downarrow$ & .3710 & .3158 $\downarrow$ & .3654 $\downarrow$ & .5599 & .5226 $\downarrow$ & .5462 $\downarrow$ \\ \bottomrule
\end{tabular}
}
\caption{Model performance on the clean, not clean (denoted as Dirty \romannumeral 1., including both \textit{input-only} and \textit{input-and-label} contaminated samples), and input-and-label contaminated (denoted as Dirty \romannumeral 2.) subsets.}
\label{tab:results}
\end{table*}

\begin{table}[t]
    \centering
    \resizebox{0.8\columnwidth}{!}{
    \begin{tabular}{lrrr}
    \toprule
         & \multicolumn{3}{c}{C-Eval}  \\
         \cmidrule(lr){2-4}
         & Clean & Dirty \romannumeral 1. & Dirty \romannumeral 2. \\
         \midrule
        Llama-2 7B & .3135 & .3344 $\uparrow$ & .3364 $\uparrow$ \\
        Mistral 7B & .4715 & .4545 $\downarrow$ & .4607 $\downarrow$ \\
        Yi 6B & .6718 & .8003 $\uparrow$ & .8117 $\uparrow$ \\
        Qwen 7B & .5619 & .6169 $\uparrow$ & .6289 $\uparrow$ \\
        Baichuan2 7B & .5508 & .5649 $\uparrow$ & .5887 $\uparrow$ \\ \midrule
        Average & .4582 & .4912 $\uparrow$ & .5012 $\uparrow$ \\
    \bottomrule
    \end{tabular}
}
    \caption{Data contamination analysis on C-Eval.}
    \label{tab:ceval-results}
\end{table}

\section{Impact of Contamination on Model Performance}
\label{impact}

To assess how data contamination impacts model evaluation, we assess over 20 popular large language models on contaminated and clean splits of each benchmark. Contaminated subsets contain examples identified in the previous section as having verbatim matches in the Common Crawl training dataset. Clean subsets on the contrary, contain examples have no matches with the training set. We employ the third party LLMs evaluation platform \textit{OpenCompass} \cite{2023opencompass} in our experiments to provide in-context demonstrations, prompts, and metrics computing. Following \cite{touvron2023llama2,openai2023gpt4}, we evaluated  Winogrande, ARC, CommonsenseQA, and HellaSwag in zero-shot setting by prompting only with the question and choices, but 5 examples are provided as context for MMLU and C-Eval. Accuracy was used as the metric. And we use perplexity to obtain the inference result, i.e., taking the choice with the lowest ppl as the predicted answer. The results are presented in Table \ref{tab:results}. Note that the Dirty \romannumeral 1. set contains both \textit{input-only} contamination and \textit{input-and-label} contamination, but the Dirty \romannumeral 2. set is solely for \textit{input-and-label} contamination. We place a $\uparrow$ for a result on dirty samples if they gain an advantage against the clean set, otherwise we add a $\downarrow$. For Llama-1,2 series models, we use the search window of 2017-2020 according to their reported training data collection period. For all other models we use an estimated search window of 2017.01-2023.10 as their exact training data collection period are unknown.

\noindent \textbf{English Benchmarks.} Based on the table, we find data contamination does not uniformly improve model performance. Instead, the impact depends on both the specific benchmark and model scale. On Hellaswag and ARC benchmarks, most models achieve better metrics on contaminated subsets. However, on MMLU tasks we observe no consistent enhancement across models. We also find that larger language models appear more capable of exploiting data contamination to achieve better performance. For instance, LLaMA-2 70B displays increased metrics on most contaminated subsets. In contrast, the 13B LLaMA-2 only outperforms on contaminated ARC. In addition, LLaMA-2 70B realises larger gains on contaminated sets (6\% and 11\% boosts on Hellaswag and ARC) than the 7B variant (5\% and 6\%). This could due to the more powerful memorisation capacity in larger language models \cite{carlini2022quantifying}. Finally, most models achieve the highest scores on the input-and-label contaminated subset versus input-only or clean sets. This proves contamination of both inputs and labels can severely affect model evaluation results. Fine-tuned models like the Llama Chat variants exhibit generally lower overall metrics compared to their foundation counterparts, but they demonstrate comparable gains on contaminated splits of Hellaswag and ARC. Specifically, the fine-tuned chat models realise similar absolute performance increases on the dirty subsets of these benchmarks as their corresponding foundation versions.

\noindent\textbf{Non-English Benchmark.} In Table \ref{tab:ceval-results}, we present contamination analysis on the non-English benchmark C-Eval. Among the tested models, Llama and Mistral are considered pure English models, while Yi, Qwen, and Baichuan are pre-trained as multilingual language models. We find the pure English models, Llama and Mistral, do not exhibit notable performance increases on C-Eval's contaminated subsets. However, the multilingual large language models all demonstrate significant performance advantages on dirty subsets. Yi 6B even achieves a 14\% higher accuracy score on the input-and-label contaminated set, proving the potential for serious distortion of evaluation results.

\begin{figure}
    \centering
    \includegraphics[width=0.85\columnwidth]{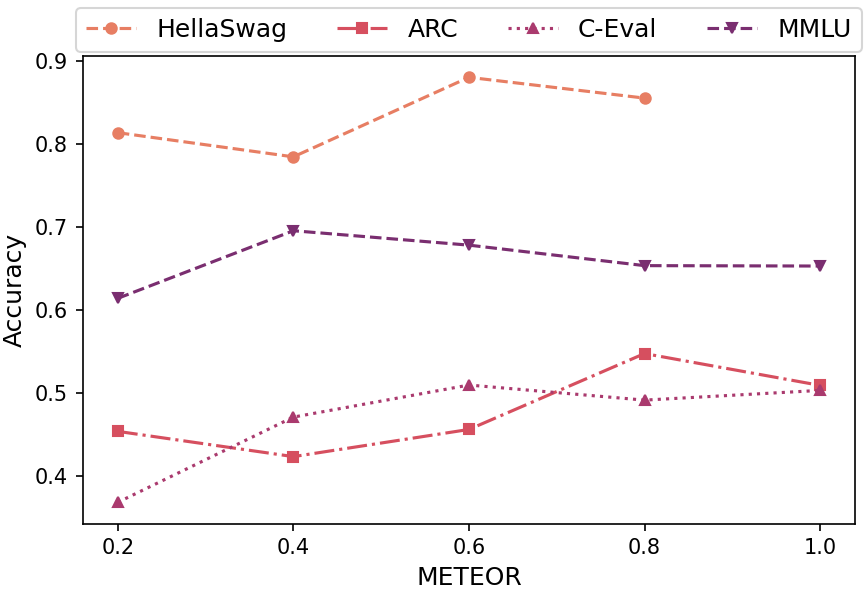}
    \caption{Accuracy of Llama-2 70B for test examples with different METEOR score. }
    \label{fig:recall}
\end{figure}

\noindent\textbf{What is the threshold of overlap for a test example to affect model prediction?} We illustrate how the METEOR score, which measures sentence similarity, correlates with model performance on test samples. The METEOR metric measures the similarity between two sentences. For instance, a test sample with a METEOR score of 0.8 indicates high equivalence between that test case and sentences in training data. In Figure \ref{fig:recall}, we group test samples by METEOR score and present the accuracy achieved on those groups by Llama-2 70B across four benchmarks. On ARC, Hellaswag, and C-Eval, a general upwards accuracy trend emerges as METEOR rises, indicating models attain higher metrics when more verbatim overlapping samples exist in the training data. In essence, substantial text duplication enables exploitation through memorisation, inflating model scores.

\section{Discussion}

\subsection{Existing Methods to Mitigate Data Contamination}

Several techniques have been proposed to mitigate data contamination issue in language model evaluation previously. Our findings provide some novel insights on the effectiveness of these approaches.

\noindent \textbf{Blocklisting benchmark sources.} 
Blocking sources of benchmarks in training data collection is a common way to avoid data contamination. In our paper, we further demonstrate the feasibility of this method. As shown in Figure \ref{fig:contamination-distribution}, the distribution of data contamination is very centralised, so blocking only a small set of domains can significantly alleviate the issue of data contamination. However, we also find blocklisted links quickly expire but content spreads, making the blocklist ineffective over time. For instance, we test the contamination blocklist in the first release of MMLU\footnote{\url{https://people.eecs.berkeley.edu/~hendrycks/data.tar}}, and we found the given blocklist only avoids 1.5\% of contaminated cases we detected in \S\ref{contamination_statistics}. If we adopt a more aggressive method that skips all domains in the blocklist, it still just avoids 21\% of contaminated cases. This suggests content used in MMLU spreads rapidly, which emphasises the necessity to update the blocklists regularly.

\noindent \textbf{Avoid using data that appears with its solution on the internet} \cite{jacovi2023stop}. According to our results, avoiding the presence of answer can indeed prevent memorising exact answers. As shown in Table \ref{tab:llama_appendix}, we found models perform generally worse on input-only contamination compare to input-and-label contamination, or sometimes even worse than the clean set. This suggest that preventing input-and-label contamination is the key to mitigate data contamination issue. However, our analysis also identify several cases where \textit{input-only} contamination provides unfair advantages. For example, most models obtain higher accuracy on the \textit{input-only} contaminated subset of ARC. As a result, completely avoiding using online resources is the best practice in benchmark construction. Winogrande is a role model in avoiding contamination by using only human-authored content.

\noindent \textbf{Protecting test data from automatic crawlers via encryption and forbidding further distribution} \cite{jacovi2023stop}. Forbidding further distribution of benchmarks can indeed prevent data contamination to some extent. This was proven in our Figure \ref{fig:contamination-distribution}, where some contaminated cases are from huggingface.co, a dataset sharing platform. However, forbidding further distribution of the test data also significantly limits the popularity of benchmarks. For example, benchmarks such as Hellaswag and C-Eval make their test sets nonpublic to avoid potential data contamination issues. However, this also makes popular third party model evaluation platforms turn to using their validation sets instead of the test sets, as the platform hosts can access the answers in the validation sets to conduct the assessment \cite{2023opencompass}. Actually, most researchers tend to evaluate their models on publicly available splits rather than restricted ones, even if the latter have lower contamination risk. Therefore, benchmarks should consider balancing robustness against ease of adoption by the community.

\subsection{Comparison to Llama's Original Contamination Report}
\label{compare to llama}

The data contamination analysis in the original Llama-2 paper is quite incomplete, presenting results for only Hellaswag and MMLU benchmarks. However, we can still compare our contamination analysis results to theirs for these two datasets. As explained in \S\ref{impact}, we use a search window of 2017-2020 for Llama-1,2 series model. In Hellaswag, we detected a similar percentage of contamination (8.3\%) to what was reported in Llama-2 (848 out of 10042 examples, 8.4\%). For MMLU, we identified a slightly lower ratio of data contamination, with 8.7\% flagged as contaminated versus 11\% marked as contaminated in the original Llama-2 paper. \citet{touvron2023llama2} showed Llama-2 70B's performance gain on HellaSwag from \textit{dirty} to \textit{not dirty}\footnote{This \textit{not dirty} set here is not the same as our clean set. See their paper for more details.} data was .0742, while our method demonstrated advantages of .0726 for \textit{input-and-label} contamination and .0622 for all contamination types. On MMLU, \citet{touvron2023llama2} only observed a performance boost from contamination on MMLU-Humanities, where Llama-2 70B had a gain (\textit{dirty} to \textit{not dirty}) of .0980 on MMLU-Humanities. In contrast, our method showed a slightly lower increase of .0845 on MMLU-Humanities (shown in Table \ref{tab:mmlu_details}). Overall, our results align well with Llama-2's original contamination reports, demonstrating the effectiveness of our methodology.

\section{Conclusion}

This paper conducted an extensive data contamination analysis for popular large language models on six multi-choice QA benchmarks. We identified varying levels of test set contamination, ranging from 1\% to 47\% across benchmarks. We also find data contamination does not necessarily lead to increased metrics: data contamination in ARC and Hellaswag generally allow models to achieve significant higher accuracy, but contamination in MMLU has little impact on model's performance. Our findings offer a transparent perspective on data contamination, emphasising its significance as an urgent issue within the evaluation community.

\section{Limitation}

Our pipeline leverages search engines followed by  querying Common Crawl index. This avoids the access of full training dataset locally, which is often not open-sourced for modern large language models. However, relying on search APIs incurs notable costs - around \$15 per 1000 queries with Bing. We spent about \$900 in total calling Bing search API. Additionally, search engines restrict query lengths, which prevents analysis of benchmarks with long input passages like reading comprehension. Moreover, large language model developers may also use customised data collected from crowd sourcing or non-public databases. In this case, our search engine plus Common Crawl pipeline may be unable to identify data contamination from these hidden data sources.

Future attempts may directly query the complete Common Crawl corpus hosted on AWS S3 services. This would enable scanning of lengthy input examples but at a higher financial cost. Alternatively, developing perplexity-based approaches to detect contaminated examples without requiring full passage matching could prove fruitful.

\bibliography{custom}
\bibliographystyle{acl_natbib}

\appendix

\section{More Information about Contamination in Multi-Choice QA Benchmarks}
\label{more-info}

To provide a straightforward impression, we provide some example of data contamination from the MMLU benchmark as shown in Figure \ref{fig:case1}. In Figure \ref{fig:case1} (a), the METEOR recall score between the test question and matched example was 0.9275, well above the 0.8 contamination threshold, indicating a clear leakage of this test example in the training data of Llama models. While minor formatting differences exist, the near-complete overlap constitutes concerning \textit{input-and-label} contamination that allows models to memorise rather than generalise. However, in Figure \ref{fig:case1} (b) we find no answer choices and the correct answer in that page, which makes it a \textit{input-only} contamination case. While \textit{input-only contamination} poses a lower risk for direct label leakage, it can still allow models unfair advantage if exposed to the questions during training.
\begin{figure}[t]
    \centering
    \includegraphics[width=0.9\columnwidth]{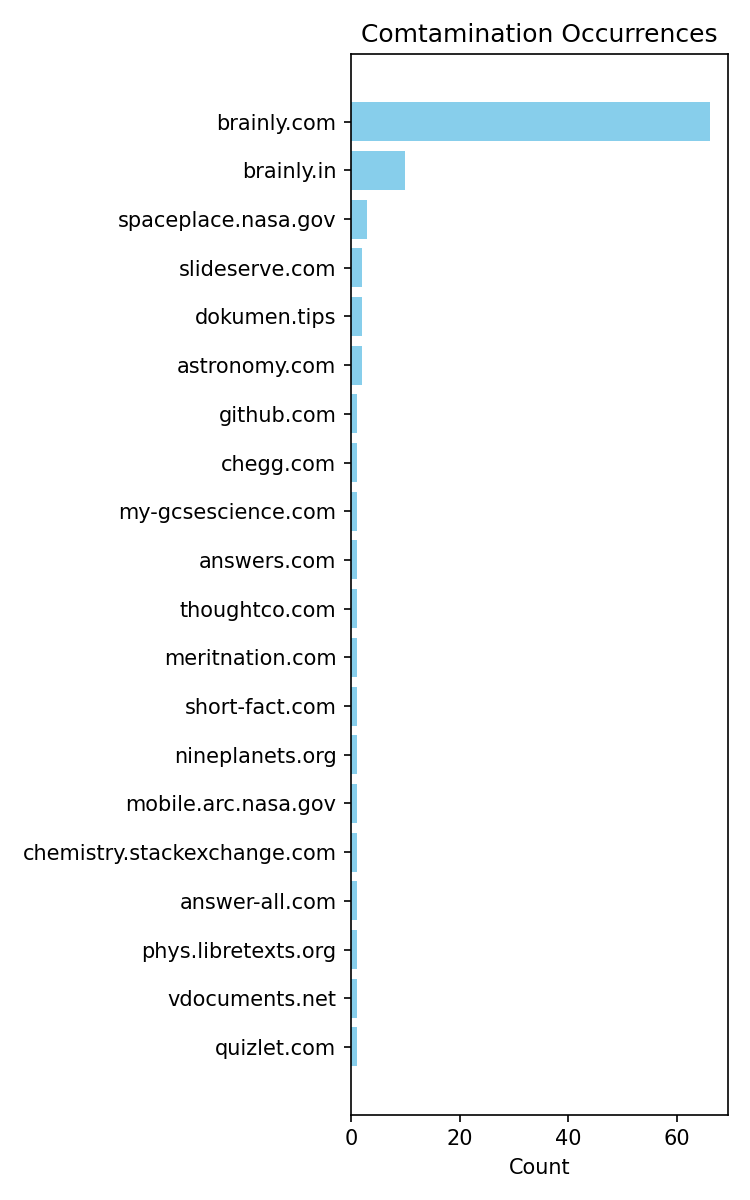}
    \caption{Domains analysis for data contamination in ARC.}
    \label{fig:arc_links}

    \centering
    \includegraphics[width=0.9\columnwidth]{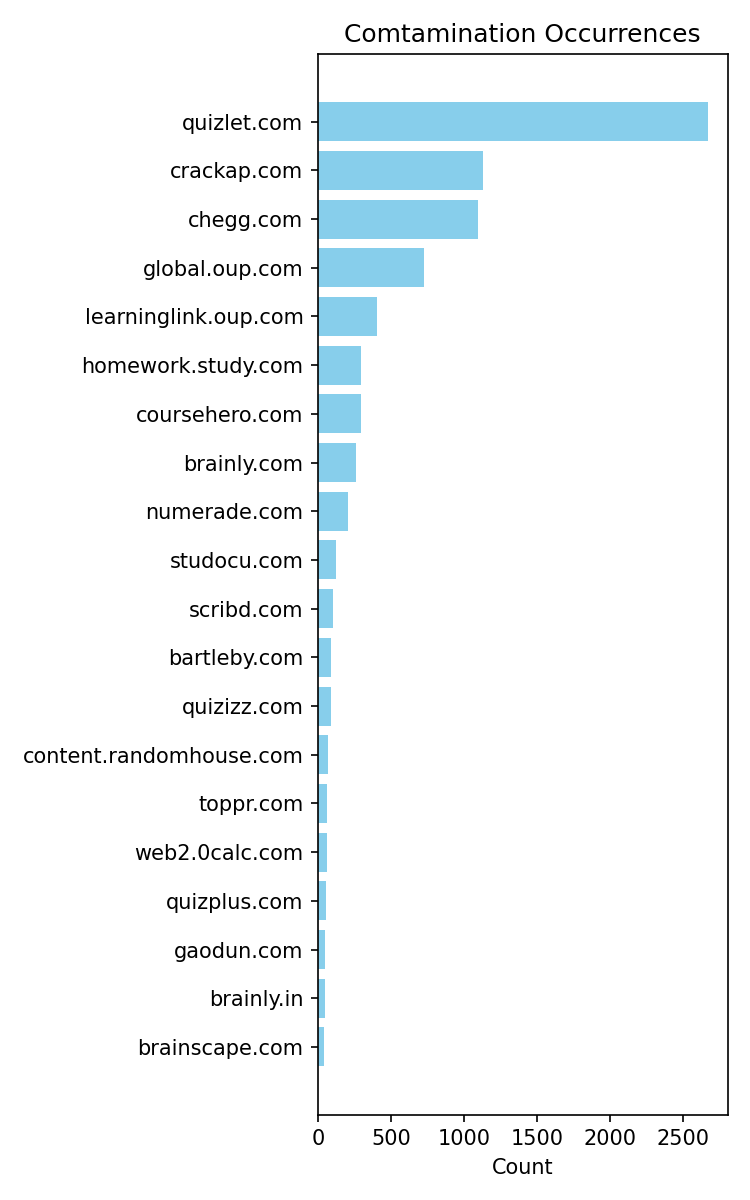}
    \caption{Domains analysis for data contamination in MMLU.}
    \label{fig:mmlu_links}
\end{figure}

We also present the domain visualisation for contaminated test sample in ARC (see Figure \ref{fig:arc_links}) and MMLU (see Figure \ref{fig:mmlu_links}).


\section{More Results on Contaminated Subsets}

We further present models' performance on the clean, input-only contaminated, and input-and-label contaminate subsets, as well as the results on the entire test set in Table \ref{tab:llama_appendix}. We found models perform generally worse on input-only contamination compare to input-and-label contamination, or sometimes even worse than the clean set. In Table \ref{tab:mmlu_details}, we present more detailed statistics of Llama models' performance on different categorises of MMLU benchmark.

\begin{table*}[ht]
\centering
\resizebox{\textwidth}{!}{
\begin{tabular}{lrrrrrrrrrrrr}
\toprule
& \multicolumn{4}{c}{MMLU} & \multicolumn{4}{c}{Hellaswag} & \multicolumn{4}{c}{ARC}\\
\cmidrule(lr){2-5}
\cmidrule(lr){6-9}
\cmidrule(lr){10-13}
& Clean & Dirty \romannumeral 1 & Dirty \romannumeral 2 & All & Clean & Dirty \romannumeral 1 & Dirty \romannumeral 2 & All & Clean & Dirty \romannumeral 1 & Dirty \romannumeral 2 & All \\
\midrule
LLaMA 7b & .3427 & .3367 & .3223 & .3356 & .7634 & .5769 & .7089 & .7560 & .3627 & .4167 & .3077 & .3614 \\
LLaMA 13b & .4652 & .3615 & .4686 & .4594 & .8347 & .7308 & .7979 & .8298 & .3930 & .3750 & .3269 & .3897 \\
LLaMA 30b & .5690 & .4563 & .5717 & .5624 & .8557 & .6923 & .8371 & .8527 & .4261 & .4167 & .4615 & .4275 \\
LLaMA 65b & .6364 & .4854 & .6525 & .6317 & .8805 & .8077 & .8612 & .8778 & .4288 & .4583 & .4615 & .4309 \\
Llama-2 7b & .4310 & .3426 & .4580 & .4340 & .7896 & .6538 & .7436 & .7834 & .3802 & .4583 & .4423 & .3845 \\
Llama-2 13b & .5647 & .4621 & .5435 & .5509 & .7236 & .6538 & .7783 & .7298 & .4233 & .4583 & .4038 & .4232 \\
Llama-2 70b & .6884 & .5671 & .7159 & .6894 & .8848 & .7692 & .8703 & .8825 & .4564 & .5417 & .5769 & .4635 \\ \midrule
Llama-2 7b Chat & .4062 & .2813 & .4060 & .3978 & .6760 & .6923 & .7632 & .6865 & .3701 & .3333 & .5000 & .3751 \\
Llama-2 13b Chat & .5417 & .4198 & .5279 & .5291 & .7341 & .6923 & .8100 & .7430 & .4334 & .5000 & .5769 & .4412 \\
Llama-2 70b Chat & .6324 & .5000 & .6392 & .6259 & .7576 & .6538 & .8341 & .7663 & .4343 & .5000 & .4615 & .4369 \\
\bottomrule
\end{tabular}
}
\caption{Llama models' performance comparison. Here Dirty \romannumeral 1. denotes input-only contamination and Dirty \romannumeral 2. demotes input-and-label contamination. All denotes the performance on the entire test set.}
\label{tab:llama_appendix}

\vspace{7mm}

    \centering
        \resizebox{\textwidth}{!}{
    \begin{tabular}{l|rrr|rrr|rrr|rrr|rrr}
        \toprule
        Model & \multicolumn{3}{c|}{MMLU} & \multicolumn{3}{c|}{MMLU-Humanities} & \multicolumn{3}{c|}{MMLU-STEM} & \multicolumn{3}{c|}{MMLU-Social-Science} & \multicolumn{3}{c}{MMLU-Other} \\
        \midrule
        & Clean & Dirty \romannumeral 1 & Dirty \romannumeral 2 & Clean & Dirty \romannumeral 1 & Dirty \romannumeral 2 & Clean & Dirty \romannumeral 1 & Dirty \romannumeral 2 & Clean & Dirty \romannumeral 1 & Dirty \romannumeral 2 & Clean & Dirty \romannumeral 1 & Dirty \romannumeral 2  \\
        \midrule
        Llama 7B  & 34.27 & 33.67 & 32.23 & 33.69 & 25.76 & 34.22 & 30.79 & 33.04 & 30.67 & 37.40 & 38.10 & 31.59 & 35.64 & 35.23 & 33.60 \\
        Llama 13B  & 46.52 & 36.15 & 46.86 & 43.79 & 43.94 & 53.38 & 37.78 & 27.73 & 37.37 & 55.55 & 49.52 & 51.33 & 50.31 & 41.48 & 49.53 \\
        Llama 30B  & 56.90 & 45.63 & 57.17 & 55.02 & 59.09 & 64.36 & 46.10 & 36.28 & 47.84 & 65.91 & 58.10 & 63.18 & 61.84 & 51.14 & 57.09 \\
        Llama 65B  & 63.64 & 48.54 & 65.25 & 63.71 & 56.06 & 74.73 & 52.58 & 41.59 & 54.09 & 72.08 & 59.05 & 74.17 & 67.30 & 52.84 & 62.21 \\
        Llama 2 7B  & 43.10 & 34.26 & 45.80 & 41.90 & 45.45 & 55.57 & 34.38 & 26.55 & 36.82 & 49.74 & 45.71 & 49.20 & 47.30 & 38.07 & 46.29 \\
        Llama 2 13B  & 56.47 & 46.21 & 54.35 & 55.73 & 59.09 & 60.91 & 44.27 & 37.17 & 44.17 & 64.22 & 60.95 & 61.69 & 62.64 & 50.00 & 54.39 \\
        Llama 2 70B  & 68.84 & 56.71 & 71.59 & 65.78 & 74.24 & 79.28 & 57.18 & 45.43 & 61.52 & 81.12 & 67.62 & 80.15 & 73.13 & 65.34 & 68.96 \\
        \bottomrule
    \end{tabular}}
    \caption{Llama series models' performance across different categories of MMLU.}
    \label{tab:mmlu_details}

\vspace{7mm}

    \centering
    \includegraphics[width=\textwidth]{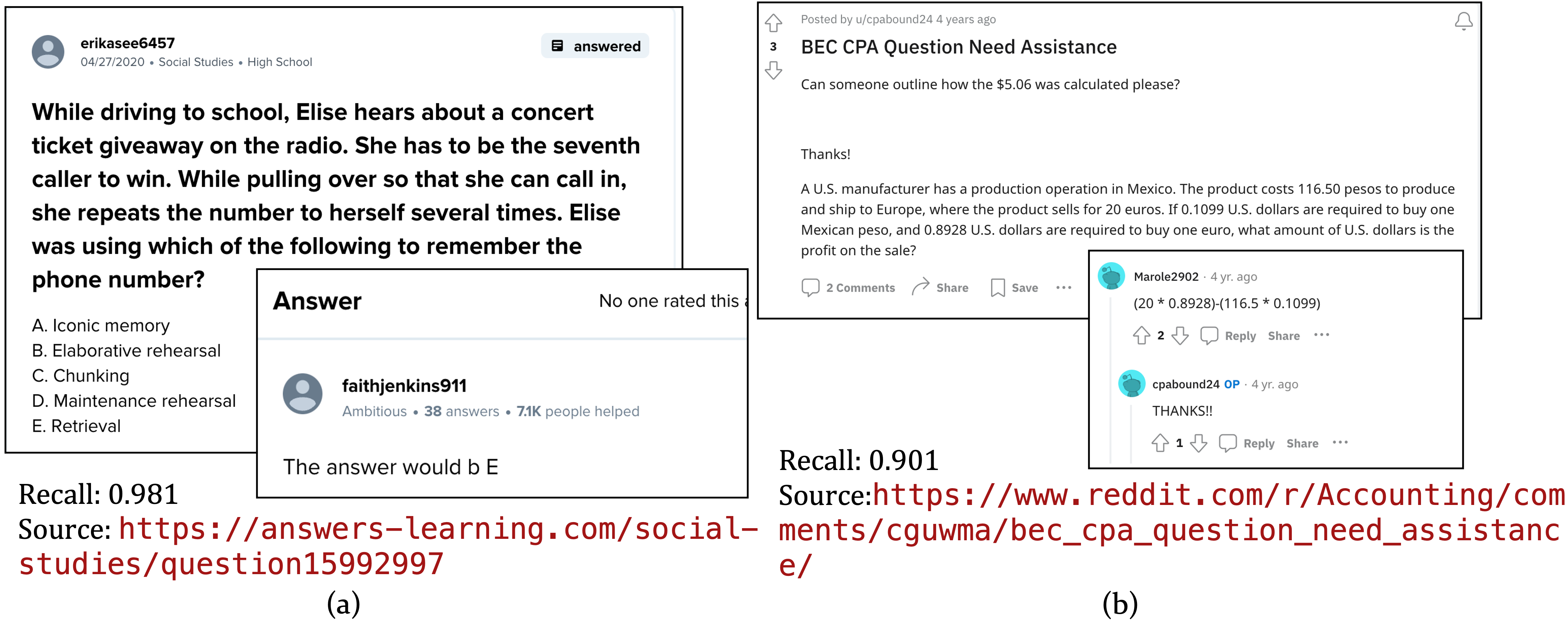}
    \captionof{figure}{An example of \textit{input-and-label} (a) and \textit{input-only} (b) contamination from MMLU.}
    \label{fig:case1}
\end{table*}

\end{document}